\documentclass{article}
\usepackage{ijcai24}

\usepackage{times}
\usepackage{soul}
\usepackage{url}
\usepackage[hidelinks]{hyperref}
\usepackage[utf8]{inputenc}
\usepackage[small]{caption}
\usepackage{graphicx}
\usepackage{amsmath}
\usepackage{amsthm}
\usepackage{booktabs}
\usepackage{algorithm}
\usepackage{algorithmic}
\usepackage[switch]{lineno}
\usepackage{multirow}
\usepackage{listings}
\usepackage{amssymb}
\usepackage{float}
\usepackage{graphicx}

\title{Human Aesthetic Preference-Based Large Text-to-Image Model Personalization: Kandinsky Generation as an Example}

\author{
Aven-Le Zhou
\and
Yu-Ao Wang \and
Wei Wu \And
Kang Zhang\\
\affiliations
Hong Kong University of Science and Technology (Guangzhou)\\
\emails
aven.le.zhou@gmail.com 
}

\begin{document}

\maketitle

\begin{figure*}[t]
    \centering
    \includegraphics[width=\textwidth]{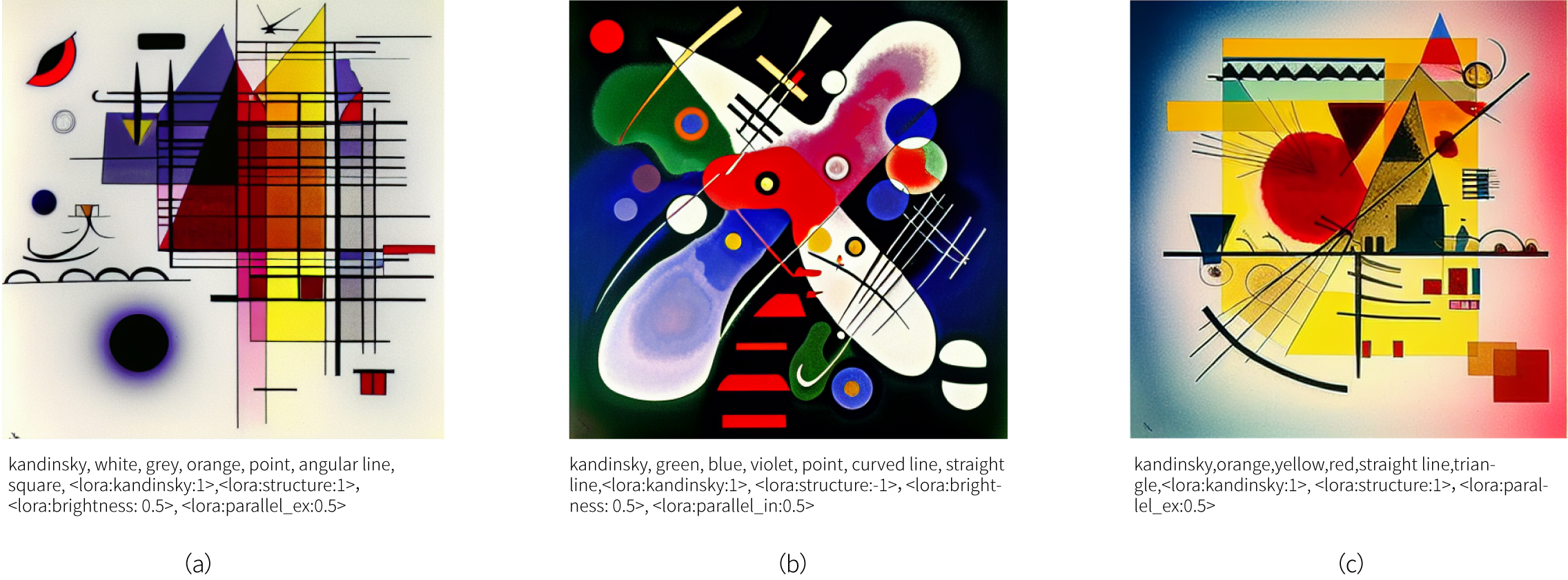}
    \caption{Artist Model: Large Text-to-Image Model Customization with Semantic Injection.}
    \label{fig:prompting}
    \end{figure*}

\begin{figure*}[t]
    \centering
    \includegraphics[width=\textwidth]{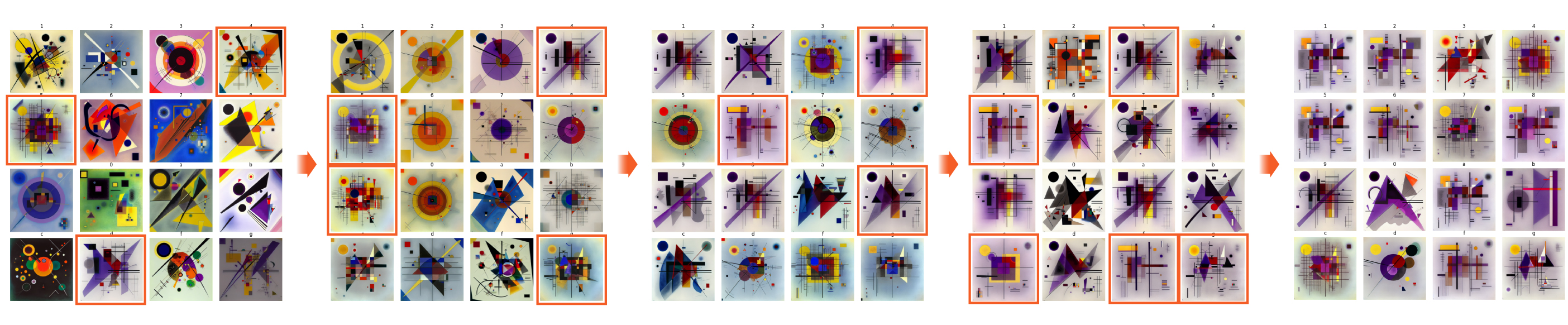}
    \caption{Prompting Model + Artist Model: Genetic Prompting Optimization with Human Feedback in Five Iterations.}
    \label{fig:demo}
    \end{figure*}

\begin{abstract}

With the advancement of neural generative capabilities, the art community has actively embraced GenAI (generative artificial intelligence) for creating painterly content. Large text-to-image models can quickly generate aesthetically pleasing outcomes. However, the process can be non-deterministic and often involves tedious trial-and-error, as users struggle with formulating effective prompts to achieve their desired results. This paper introduces a prompting-free generative approach that empowers users to automatically generate personalized painterly content that incorporates their aesthetic preferences in a customized artistic style. This approach involves utilizing ``semantic injection'' to customize an artist model in a specific artistic style, and further leveraging a genetic algorithm to optimize the prompt generation process through real-time iterative human feedback. By solely relying on the user's aesthetic evaluation and preference for the artist model-generated images, this approach creates the user a personalized model that encompasses their aesthetic preferences and the customized artistic style.
\end{abstract}

\section{Introduction}

Generative Artificial Intelligence (GenAI) has been developed for various forms of content, including text, sounds, images and videos, demonstrating its potentials of facilitating human creativity \cite{dalle,Rombach_Blattmann_Lorenz_Esser_Ommer_2022,text2image_diffusion,yu2022scaling,beyond_prompt}. One of the most promising applications is large text-to-image models \cite{best_prompt}. Several text-guided image generation solutions, such as DALL-E 2 \cite{dalle} and Stable Diffusion \cite{Rombach_Blattmann_Lorenz_Esser_Ommer_2022}, utilize joint image and text embedding learning (i.e., CLIP \cite{clip}) and diffusion models \cite{diffusion_model} to produce realistic and aesthetically-pleasing contents. These models have become popular for creating digital images \cite{VQGAN} and artworks \cite{Rombach_Blattmann_Ommer_2022,creativity_of_text2image}.

Recently, the text-to-image art (TTI-art) community has actively engaged in utilizing a variety of available resources, such as the Stable Diffusion Web UI \footnote{\url{https://github.com/AUTOMATIC1111/stable-diffusion-webui}} and commercial products like Midjourney \footnote{\url{https://www.midjourney.com/}}. This community comprises both amateurs and professionals, including novice and skilled artists. Given the growing capabilities and and user-friendly nature of text-to-image generation ecosystem, digital art created using these large models has experienced significant progress and is on the brink of becoming a mainstream phenomenon \cite{creativity_of_text2image}.

\subsection{Prompting in Text-to-Image Generation}

Common in text-to-image generation systems is the ability for users to create digital images and artworks by writing prompts in natural language. This process, known as prompt engineering \cite{prompt_engineering}, prompt programming \cite{prompt_programming}, prompt design \cite{prompt_design}, or simply prompting \cite{creativity_of_text2image}, allows the user to generate content without requiring in-depth knowledge of the underlying technologies. However, it is important to note that the interaction between the user and AI are not always centered around the human experience  \cite{multi_concept}. When creating art from text, the user relinquish a certain degree of control to AI \cite{Galanter_2016}.

The interaction between humans and AI in this case restricts the input to a specific paradigm that is more suitable for the models rather than considering the needs and intuition of the human user \cite{multi_concept}.  Consequently, the prompts generated under this paradigm often appear arbitrary \cite{best_prompt}. It is remarkable that large text-to-image models are capable of easily generating aesthetically-appealing results. Due to the unstructured and open-ended nature of text inputs, however, the process often becomes a brute-force trial and error  \cite{prompt_engineering}. Users must continuously search for new text prompts to iterate on their preferred image generation, which is ``random and unprincipled'' \cite{prompt_engineering}. In other words, the process can be non-deterministic; prompting remains an ongoing challenge for end users.

\subsection{Controlling Non-determinism}

Chaos in the context of digital art, particularly in generative art such as randomness and emergence  \cite{art_emergence_random,emergence_genart}, is not always viewed as a negative phenomenon but rather as opportunities that artists prefer to work with \cite{ten_questions}. Generative artists employ processes that possess a certain degree of autonomy to create all or part of an artwork \cite{Galanter_2016}. Art programmed using a computer intentionally introduces chaos, such as randomness, as part of its creation process \cite{Bailey_2018}. Artists hope that emergence occurs, resulting in interesting and unforeseen outcomes surpassing initial expectations \cite{emergence_genart}. The exploration of non-deterministic artworks in a controlled manner and finding the ``sweet spot'' between full control and total chaos is not a new concept. Benjamin Kovach describes generative art as a continual search for the ideal balance between complete control and total chaos \cite{Kovach_2018}.

When confronted with non-deterministic text-to-image models, it is important to consider how generative art can contribute to the development of such AI technology and the role of artists in this process. We argue that understanding and leveraging generative art experiences can offer fresh insights into how one can better collaborate with GenAI. To substantiate the control of non-determinism in GenAI from an artistic standpoint, we explore how ``traditional'' generative art approaches like procedural modeling and genetic algorithmic (GA) generation can be applied to large text-to-image models for artistic painterly content generation.

\subsection{Customizing Large Text-to-Image Model}

Large models possess unparalleled capabilities in high-quality and diverse synthesis of images and text, thanks to the strong semantic knowledge acquired from a vast collection of image-caption pairs \cite{dalle,Barron_2021}. These models operate in a universal and general manner, lacking the ability to generate results tailored to the user's preferences within a given reference set \cite{dreambooth}. Even the most detailed description may produce undesired appearances since the text embedding space cannot precisely reconstruct the desired preference, only generating variations \cite{clip}. 

Low-Rank Adaptation of Large Language Models (LoRA) \cite{Hu_lora} has sparked numerous derivative techniques and applications to customize large text-to-image models in TTI-art community. Considering the popularity of GenAI as a new artistic medium and the tremendous efforts in the community, the experience and techniques derived from art can prove to be beneficial. We adapt experiences from the TTI-art community, to incorporate several techniques from the community to customize a large text-to-image model with given artist style, referred to as ``artist model'' (in Fig. \ref{fig:prompting}).  

\subsection{Our Prompting-Free Approach}
To inherit generative art experiences in evolutionary generation, we propose to apply GA technique with real-time iterative human feedback, to optimize prompting automatically and generate customized images using our ``artist model.'' Initially, we establish a semantic descriptive guideline (of given artist style), base on which we design a procedural prompting model and then apply evolutionary optimization on it with real-time iterative feedback from the user, relying solely on the user's aesthetic evaluation and preference for the resulting images. Ultimately, we obtain an optimized prompting model incorporated with user-preference (see Fig. \ref{fig:demo}). The optimized prompting model can automatically generate user-preferred prompts without the need for further input. 

By combining the artist model with the optimized prompting model, we provide an optimized generative solution that produces a personalized model incorporating customized artistic style and user's preference.The personalized model require no further input but automatically generates based on the user's preference, it is prompting free! Our contributions can be summarized as follows:


\begin{enumerate}
    
    \item A generative approach empowering the user to automatically create personalized painterly content based on their aesthetic preferences in customized artistic style, eliminating the need for explicit prompting. This approach involves ``semantic injection'' to customize the large text-to-image model and genetic prompting optimization with real-time iterative feedback. 
        
    \item Construction of a text-to-image dataset of Wassily Kandinsky and validation of the above techniques and algorithms as an artistic experiment. The codes and dataset are open-sourced to the public.
    
\end{enumerate}

\section{Related Work}
\subsection{Text-to-Image Models Customization}

Several studies have explored techniques for customizing large text-to-image models to specific object. DreamBooth customizes text-to-image diffusion models by fine-tuning them with a small number of images \cite{dreambooth}. This approach enables the synthesis of photo-realistic images while preserving the key features. Similarly, \cite{Roich_Mokady_2021} use textual inversion method to represent new concepts (i.e., a few user-provided images) through new ``words'' in its embedding space. \cite{Wei_Zuo_2023}  introduce a learning-based encoder that enables fast and accurate customized text-to-image generation. The low-rank adaptation (LoRA) method reduces the number of trainable parameters for downstream tasks by freezing pretrained model weights \cite{Hu_lora}. \cite{Feng_He_Fu_2023} propose a training-free method for guiding diffusion models to improve attribute-binding and compositional capabilities. Furthermore, \cite{Wolf_2023} introduce another training-free strategy called FABRIC, which utilizes the self-attention layer in diffusion models to optimize customized content creation.

\subsection{Human Preference and Feedback}
Various studies have focused on modeling human preferences and leveraging human feedback in text-to-image research in different ways. \cite{Wu_Sun_Zhu_Zhao_Li_2023} propose a method to address the issue of text-to-image models not aligning well with human preferences by utilizing a human preference classifier to adapt the model and generate preferred images. \cite{Tang_Rybin_Chang_2023} present ZO-RankSGD algorithm improves the quality of images generated by a diffusion generative model with human ranking feedback. \cite{Hao_Chi_Dong_Wei_2023} propose a prompt adaptation framework that enhances text-to-image models by automatically taking user input and adapting to the model-preferred prompt. Many alike works utilize human preference and feedback to improve large text-to-image performance, but less to solve user's prompting problem or generate user-preferred prompts. \cite{best_prompt} proposes a human-in-the-loop approach that employs a genetic algorithm to identify the most optimal combination of prompt keywords. This work shares one similar method with ours, although it focuses exclusively on searching for the best prompt to facilitate the use of large-scale models in general.

\section{Method}



\subsection{Semantic Injection}
    
Building upon open-source research efforts like LoRA, the TTI-art community has dedicated substantial efforts and exploration, resulting in the development of several effective variations. Our research delves into these community efforts and adapts several effective techniques to propose the concept of ``semantic injection,'' which incorporates techniques like fast LoRA \footnote{\url{https://github.com/cloneofsimo/lora}} and DiffLoRA \footnote{\url{https://pixai.art/model/1630413070130926806}} to customize the large model. 

\paragraph*{Fast LoRA}

Fast LoRA technique aims to balance between the quality and complexity of fine-tuning stable diffusion in time and space. Built upon research efforts such as pivotal tuning \cite{Roich_Mokady_2021}, DreamBooth \cite{dreambooth} and textual inversion \cite{Gal_Alaluf_2022}, fast LoRA technique adjusts transformer model's attention layers by adding low-rank matrices to the model's weights to achieve significant changes in model behavior by modifying a small number of parameters. For the pre-trained weight matrix $W \in \mathbb{R}^{n\times m}$, the adjusted weight matrix can be represented as $W' = W + \Delta W = W+AB^{T}$, where $A\in \mathbb{R}^{n\times d}$ and $B\in \mathbb{R}^{m\times d}$, $n$ is the dimension of the original weight matrix and $d$ is the low-rank factor and smaller than $d$. Notably, residual $\Delta W$ is decomposed into the product of low-rank matrices $A$ and $B$ with much smaller size, allowing efficient tuning of the model instead of updating larger $W$. Furthermore, this technique only fine-tune parts of the transformer model (i.e., self-attention heads), to reduce final model size. 

\paragraph*{DiffLoRA}
DiffLoRA technique is a multi-step operation that utilizes LoRA and two images: original image A and manipulated image B. Initially, we fine-tune original stable diffusion model with the image A using LoRA until it over-fits, resulting in Lora A. We then combine this over-fitting Lora A with the original model to create model A. Subsequently, we repeat the same process using the model A (not the original model) and the image B, generating the over-fitting LoRA B and model B. Finally, we leverage the discrepancy between the model B and the original model to establish DiffLoRA. This technique effectively captures and learns the critical differences between the two inputs as the desired attribute range of continuous values.
    
Combining these techniques, semantic injection can efficiently encode the description of discrete and continuous values of different attributes into the large model. The refinement over large model can acquire understanding and effective synthesis of given artistic style (with its semantic description). Incorporating the various fine-tuned LoRAs with the original model as a customized model, semantic injection leads to the \textbf{artist model} with given artistic style.


\subsection{Genetic Prompting Optimization}

The genetic algorithm (GA), as an evolutionary algorithm, mimics the natural evolutionary process through selection, crossover, and mutation operations on individuals \cite{Holland_1992}. With the customized artist model, we define the acquisition of a personalized model as a prompt optimization problem. To address this problem, we propose to apply GA with real-time iterative human feedback for evolutionary optimization. We proceed with the artist model and define the personalized generative model as $G: C \to I$, where $C$ represents chromosomes indicating prompts, and $I$ represents individuals in the population which are the generated images. 

\begin{figure}[h]
    \centering
    \includegraphics[width= 0.475\textwidth]{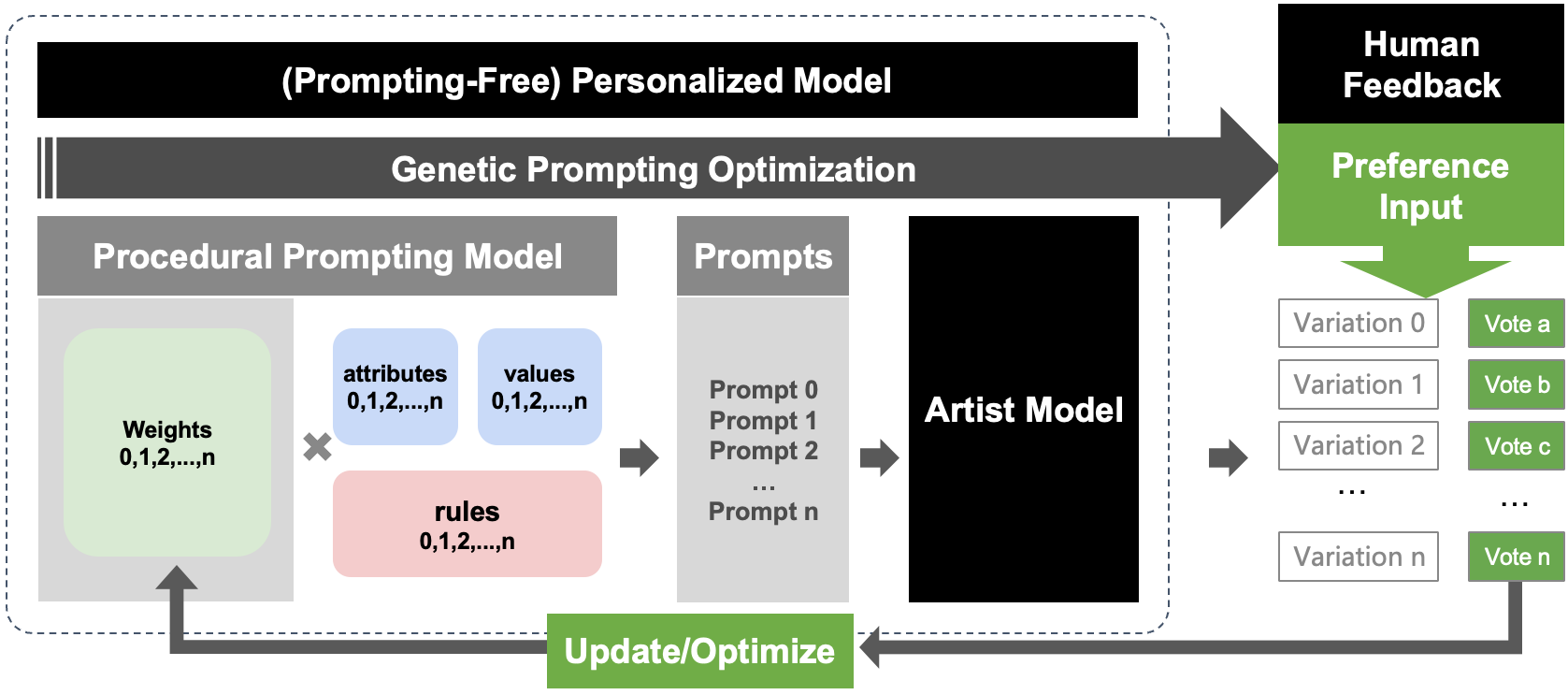}
    \caption{Genetic Prompting Optimization with Human Feedback.}
    \label{fig:genetic_optimization}
    \end{figure}

In our approach, the individuals represent images generated by the artist model, while the chromosomes represent the user's prompts consist of genes representing the attributes of given artistic style. We assume that the user maintain a consistent tendency in their aesthetic preferences during the optimization. The fitness of the individuals is evaluated based on the user's aesthetic preferences in each iteration. Following Darwin's theory of ``survival of the fittest'' \cite{Grefenstette_1993}, the prompts that generate images with higher fitness become survivors after several iterations of evolution, leading to a optimized prompting model incorporated with user's preference. Fig. \ref{fig:genetic_optimization} shows the process in details. The optimized prompting model can automatically generate preferred prompt set without the need for further input from the user. This method allows the user to obtain an automatic prompting model, relying solely on their aesthetic evaluation and preference. 

\section{Kandinsky Bauhaus Style}

Wassily Kandinsky endeavors to evoke emotional responses through the use of color, form, and composition \cite{Johnston_2012}, and abandons recognizable imagery and relying solely on color and non-representational forms to depict the ``invisible reality,'' thus becoming pioneers in the evolution towards “consciousness for humanity” \cite{Walker_2014}. Kandinsky's artistic style reached a prominent phase upon his return to the Bauhaus in 1922. This period (1922-1933) also marked the publication of his significant work, ``Point and Line to Plane,'' in which Kandinsky has thoroughly explained his art theory about color, form and composition. Along with his excellent corpus of paintings executed under the guidance of his theory, Kandinsky's Bauhaus style becomes a perfect yet challenging artistic style to implement our proposed approach.

\begin{figure}[h]
    \centering
    \includegraphics[width=0.48\textwidth]{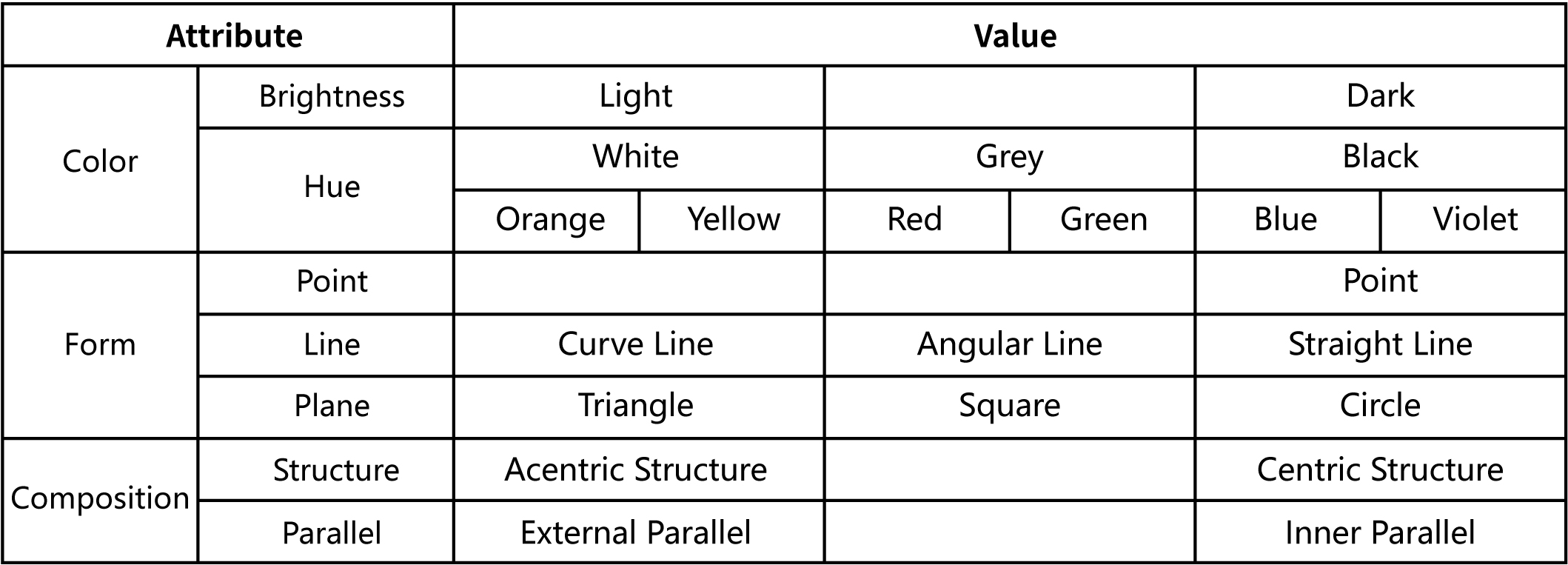}
    \caption{Semantic Descriptive Guideline: Attribute-Values.}
    \label{fig:attribute-value}
    \end{figure}

\subsection{Kandinsky Semantic Descriptive Guideline} 
To systematically describe Kandinsky's Bauhaus style, we collaborate with researchers specializing in Kandinsky's art and examine the texts in ``Concerning the Spiritual in Art'' \cite{Kandinsky_1977} and ``Point and Line to Plane'' \cite{Kandinsky_Rebay_1979} to establish a semantic descriptive guideline.  Due to page limitation, we include our analysis of Kandinsky's color, form, and composition theory very breifly in appendix \ref*{cfc} as supplementary material. We summarize the literature and Kandinsky theory and derive a list of attributes and values that encompass Kandinsky's style. Fig. \ref*{fig:attribute-value} summarizes the list which consists of 7 attributes and 22 values.


\subsection{Kandinsky Bauhaus Style Paintings}

We gather 209 Kandinsky paintings during his Bauhaus period and meticulously narrow down the selection to 65 representative artworks through expert evaluation. The selection criteria encompasses two key aspects: (1) the artworks need to be complete pieces rather than incomplete sketches, and (2) they are required to exhibit explicit attributes that align with the descriptions presented in ``Point and Line to Plane'' and our semantic descriptive guideline. 

We collaborate with experts to categorize and label the selected paintings using our semantic descriptive guideline. The labels start with a prefix of ``Kandinsky'' and are followed by the attribute values. We selectively retain the attribute values that have discernible effects and omit those that are non-obvious or controversial. This empirical approach ensures a well-balanced dataset across various attributes, encompassing distinct and explicit Kandinsky features. The dataset is open-sourced to the public.


\section{Experiment I: Artist Text-to-Image Model}
Stable Diffusion is one of the most popular open-source solutions for text-to-image generation due to its powerful features and user-friendly interface \cite{Rombach_Blattmann_Lorenz_Esser_Ommer_2022}. Similar as most large models, it is not built for specific artistic styles, such as Kandinsky. We utilize the standard Stable Diffusion model \footnote{\url{https://huggingface.co/runwayml/stable-diffusion-v1-5}} and employ two fine-tuning techniques (i.e., fast LoRA and DiffLoRA from the TTI-art community. 

To semantically inject discrete values like form-related attributes such as point, line, and plane, as well as color hue, we employ the fast LoRA technique. Although color hue is typically considered a continuous value, Kandinsky defines it as six primary colors, thus we use it as discrete values. For color brightness and composition-related attributes, we utilize DiffLoRA to derive three additional LoRA models. By combining these four LoRAs, we complete the ``semantic injection'' of Kandinsky's attribute-value list (see Fig. \ref{fig:attribute-value}). Furthermore, we construct a baseline model to facilitate a comparative evaluation of the proposed method.

\begin{figure}[h]
    \centering
    \includegraphics[width=0.475\textwidth]{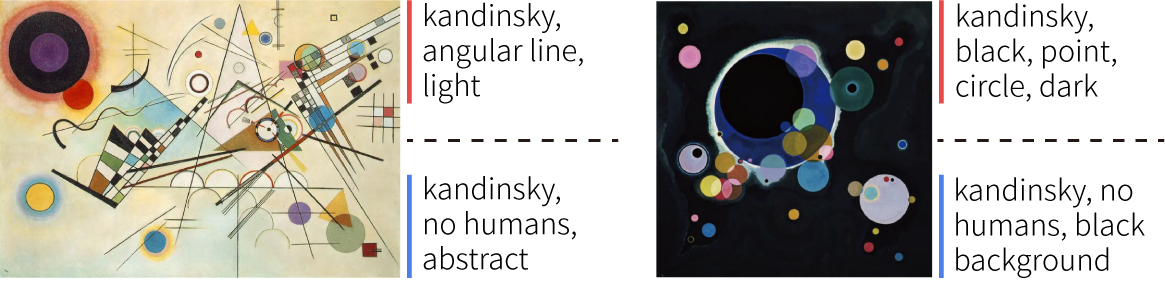}
    \caption{Kandinsky Text-to-Image Dataset.}
    \label{fig:dataset}
    \end{figure}
    
\subsection{Fast Lora Experiment}

We initially employ an automatic labeling extension \footnote{\url{https://github.com/toriato/stable-diffusion-webui-wd14-tagger}} from the community to annotate the collected Kandinsky paintings. Fig. \ref{fig:dataset} shows an example of Kandinsky paintings and descriptions in text-image pairs, blue labels are auto-generated while red labels are from our dataset. We utilize the fast LoRA technique to fine-tune the stable diffusion model using the automatically labeled data and our Kandinsky dataset (hue and form-related attributes only), resulting in the baseline model and our the fine-tuned fast LoRA model, respectively. 

\begin{figure*}[!htb]
    \centering
    \includegraphics[width=\textwidth]{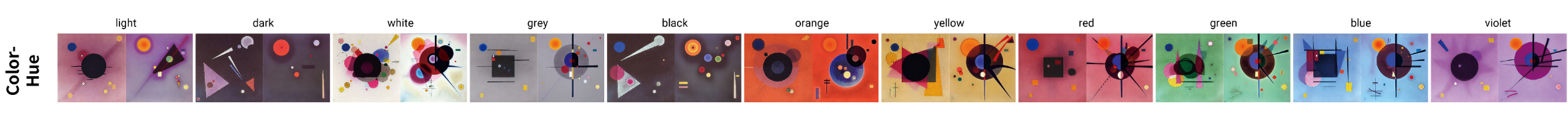}
    \caption{Baseline Model Performance on Hue Attribute.}
    \label{fig:hue_result}
    \end{figure*}

\begin{figure*}[!htb]
    \centering
    \includegraphics[width=\textwidth]{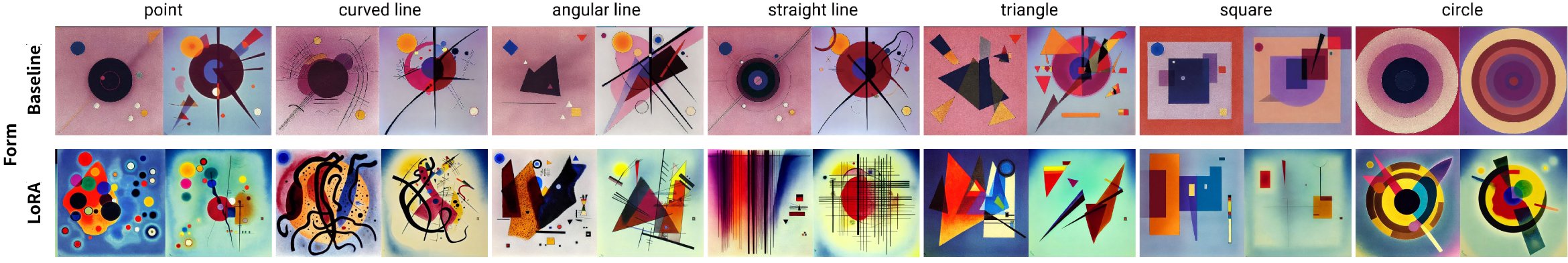}
    \caption{Fast LoRA vs. Baseline on Forms-Related Attributes.}
    \label{fig:fastlora_result}
    \end{figure*}

For evaluation, we utilize attribute value as single-keyword prompts to generate images using both models. Fig. \ref{fig:hue_result} demonstrates the effective performance of the baseline model on the color hue attribute, leading us to consider the color hue attribute as inherent to the original model. Due to space limitations, we chose to omit the similar results produced by the fine-tuned fast LoRA model. Fig. \ref*{fig:fastlora_result} showcases the comparative results of the baseline model and fast LoRA on form-related attributes, highlighting the significantly more effective performance of fast LoRA in generating the corresponding Kandinsky features.

\begin{figure}[h]
    \centering
    \includegraphics[width=0.475\textwidth]{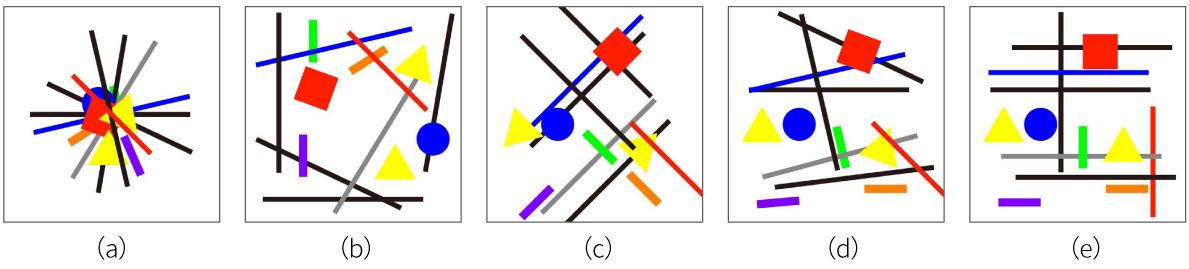}
    \caption{Images for DiffLoRA Fine-tuning.}
    \label{fig:difflora_data}
    \end{figure}
    
\subsection{DiffLoRA Experiment}

\begin{figure*}[!htb]
    \centering
    \includegraphics[width=\textwidth]{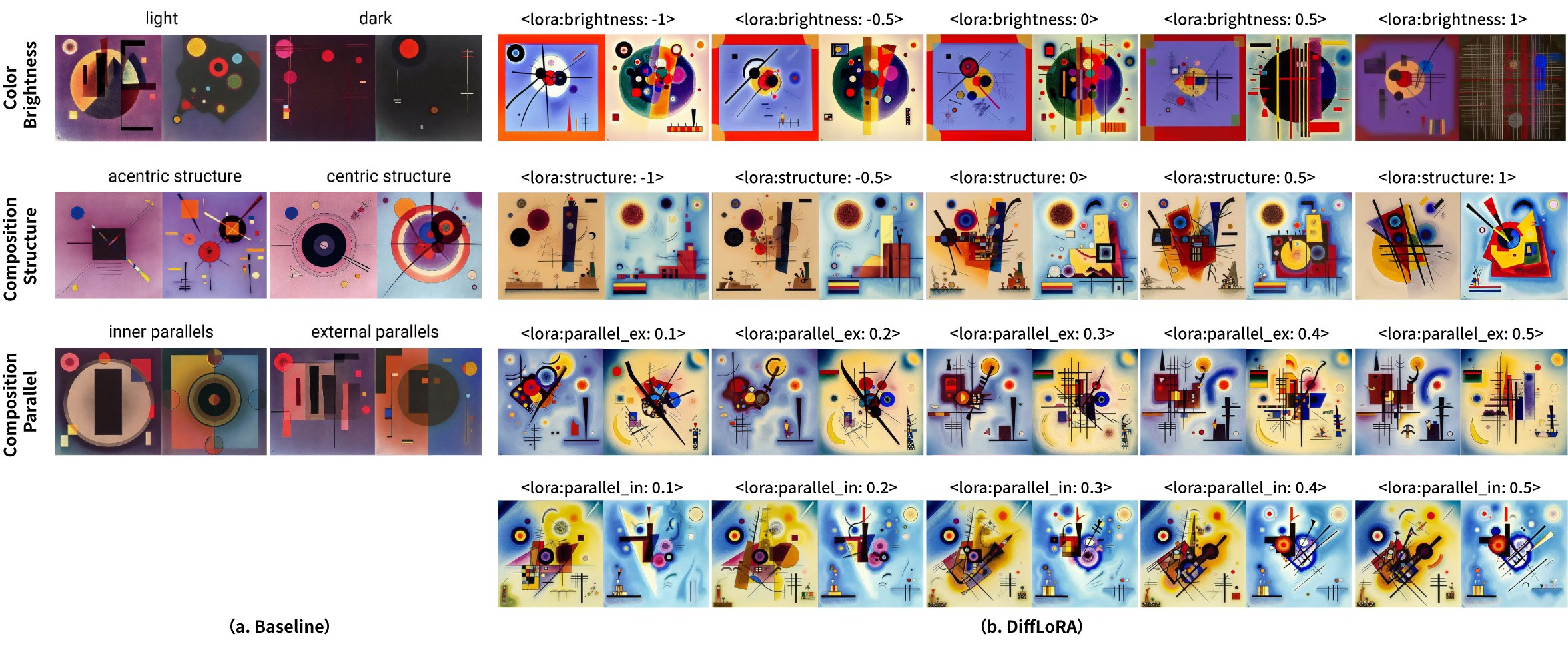}
    \caption{DiffLoRA vs. Baseline on Brightness and Composition-Related Attributes.}
    \label{fig:difflora_result}
    \end{figure*}

To incorporate the brightness attribute, we choose a Kandinsky painting as the reference image A and adjust its brightness to create a much darker image B, and then generate the DiffLoRA for brightness. For the structure and parallel composition attributes, modifying an existing painting to reflect these features is challenging. Therefore, we manually create the inputs to represent these attributes (see Fig. \ref*{fig:difflora_data}). Kandinsky's definitions of inner parallel elements as diagonal and external parallel elements as edges deviate from conventional understanding, making it difficult for the model to learn this attribute. To address this, we introduce Fig. \ref*{fig:difflora_data} (d) as an intermediate input of parallel structure and pair it with two extreme cases (i.e., Fig. \ref*{fig:difflora_data} (c) \& (e) for inner and external parallel respectively), to create two DiffLoRA models that effectively represent this attribute. Fig. \ref*{fig:difflora_result} demonstrates that the DiffLoRA can synthesize corresponding Kandinsky features based on single-attribute prompting. 

\begin{figure}[h]
    \centering
    \includegraphics[width=0.475\textwidth]{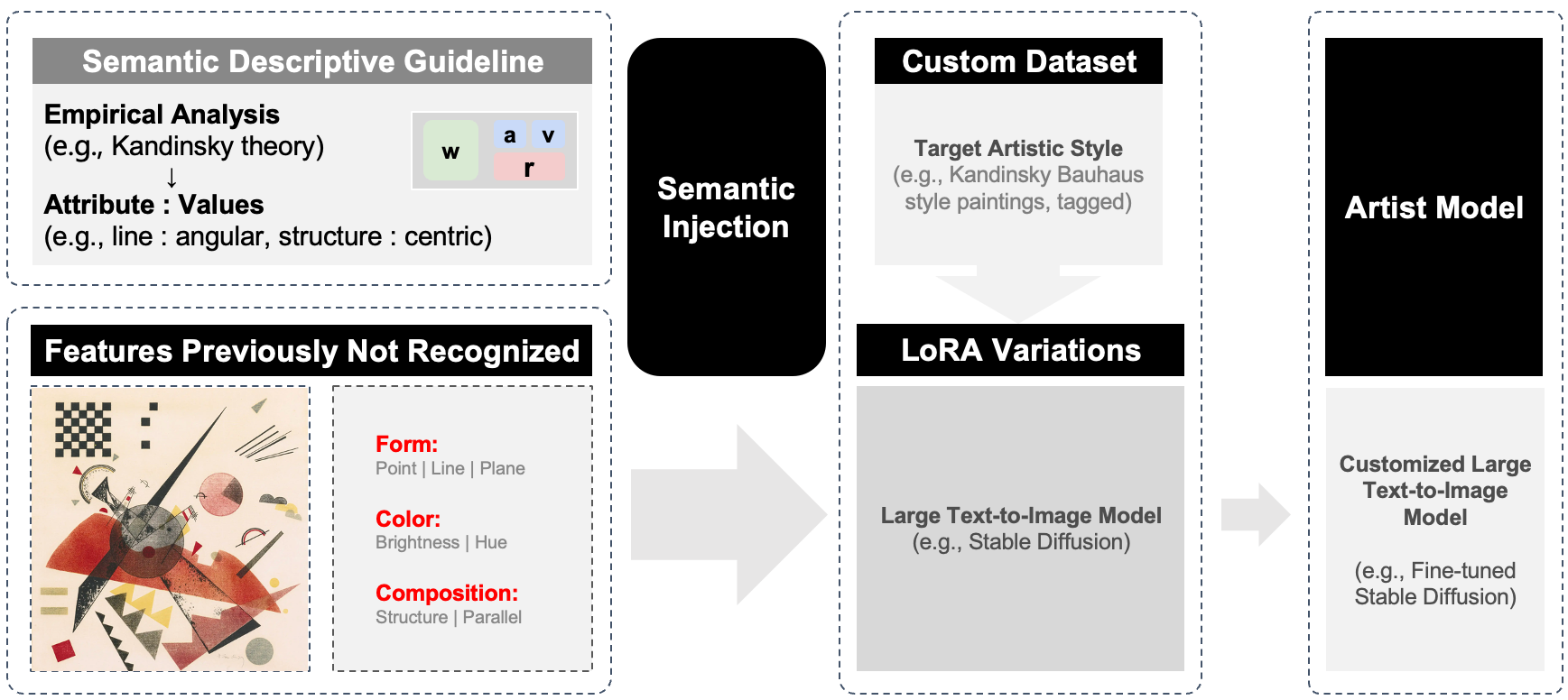}
    \caption{Artist Model Customization with Semantic Injection.}
    \label{fig:artist_model} 
    \end{figure}
    
\subsection{Artist Model and Result}

Fig. \ref{fig:artist_model} depicts the detail process of obtaining our artist model. We incorporate the fine-tuned fast LoRA and DiffLoRA to create the customized model, i.e., our artist model. We collaborate with the Kandinsky expert to validate its generative capability. We test with human-created prompts refer to our semantic descriptive guideline to generate Kandinsky-style artworks. For example, Fig. \ref{fig:prompting} (a) show the prompt and the generated image which the expert is prompting a painting with \textit{``warm-temperature colors (e.g., orange), light tones (e.g., white, grey), and neutral bright color elements (such as points, angular lines, and squares) arranged parallel to the edges of the painting in an acentric structure.''}

Fig. \ref{fig:prompting} (b) and (c) display two additional examples. If one possesses an understanding of our semantic descriptive guideline or theory of Kandinsky's Bauhaus style, they can utilize the customized artist model and manually prompting to generate his Bauhaus-style paintings according to their preference. However, this semi-controllable process still requires prompting and does not yet align with our proposed prompting-free appraoch.


\section{Experiment II: Prompting-Free Personalized Model}
Based on our semantic descriptive guideline and attribute-values of Kandinsky Bauhaus style, we implement the genetic prompting optimization method in following steps:

\begin{enumerate}

\item Initial Generation: The procedural model constructed from the semantic descriptive guideline generates prompts through a procedural process and the artist model generates images accordingly, resulting with the initial population consists of $n$ individuals.

\item  Voting: The user votes for their preferred individuals within the population. The fitness function evaluates the individuals' fitness (based on the user's inputs) and selects survivors (along with their corresponding chromosomes, i.e., prompts) for the next generation.

\item  Evolution: The genetic mechanism produces new population of individuals using the selected survivors and repeats this process until the user is satisfied with the results, at which point the prompting model is optimized.

\end{enumerate}

\subsection{Procedural Prompting Model}

The prompting of our artist model encompasses attribute-value pairs (hue, point, line, and plane) and the names of the DiffLoRA models with relevant parameters (brightness, structure, and parallel) pertaining to color, form, and composition (see Fig. \ref*{fig:attribute-value} for the complete list and Fig. \ref*{fig:prompting} for examples). While brightness, structure, and parallel are continuous-value attributes, the other attributes are discrete values. For discrete-value attributes, we establish the following prompting rules. Hue attribute selects three values and line selects a single value. Since the point attribute has only one value, we combine it with plane and let it select two from the four values. 

The attributes with continuous values are denoted as $A_C$, the attributes consisting of a single discrete value are represented by $A_S$, and $A_M$ represents the attributes consisting of multiple discrete values. Due to the distinct types of these attribute-values, we employ Heterogeneous Encoding \cite{Fall_1995} to represent the information in prompts as genes. The chromosome is described as $C={\left\{Style,A_{C},A_{S},A_{M},S \right\}}$. $C$ represent gene positions for different attributes divided by the type of value of attributes. $Style$ represents the consistent keyword in the prompt, i.e., Kandinsky in this case. $Seed$ represents the random seed of Stable Diffusion. In our experiment, we initialize a set of seeds at the first generation by assigning an integer in $[0,2147483647)$.

\subsection{Genetic Prompting Optimization}

\paragraph*{Fitness Function} 

Our approach incorporates real-time human feedback (i.e., the user's votes) to assess the fitness of individuals. In each iteration, the user has the opportunity to vote for the individuals within a population of $n=16$. The number of votes received by an individual is denoted as $V_i$, where $i\in(1,16)$ and $V_i\in(0,+\infty)$. The fitness function is defined as $f(i)=V_i$. We assign weights to discrete values of $A_S$ and $A_M$ for optimization. 

\paragraph*{Weight Updating} 
Initially, all possible values have weights set to 1, denoted as $w_{v}=1$, $v$ represents one attribute-value in the prompt. In each iteration, the updated weight $w_{v}'=w_{v}+\sum^{n}_{i=0}V_i$, where $v \in C_{i}$,  $C$ is the chromosome (i.e., prompt) for individual $i$. Based on the assumption of a consistent tendency of user preference, we assume the continuous-value attribute should adhere to a normal distribution and incorporate user inputs to update the distribution (i.e., mean and variance), to reflect the evolving characteristics of dominant individuals within the population. 

\paragraph*{Selection} 
We facilitate the Roulette Wheel Selection \cite{Holland_1992} in our experiment and calculate the probability of chosen $p(i)$, which is defined as $P_i =\frac{F_i}{\sum^{n}_{j=1}F_j}$, where $n$ is the number of individuals in population. Due to the small-scale population size ($n=16$), two parents are selected for one individual in the next generation. 

\paragraph*{Crossover} 
To combine genetic information from the two parents and produce offspring, we utilize different strategies for the various types of genes. For $A_S$ and $Seed$ in the offspring's chromosome, we employ Uniform Crossover \cite{Leong_2017} with a fixed probability of $p=0.5$ to determine which parent's genes will be inherited. For $A_M$, we conduct a without-replacement drawing from the set of possible values. As for the gene position $A_C$, we generate the offspring's value by applying average crossover on values of the parents.

\paragraph*{Mutation} 
We employ the Uniform Mutation technique for attributes with discrete values, i.e., $A_S,A_M,Seed$ and set the mutation rate as $p=0.05$. An attribute's value is modified by randomly selecting a new value from its set of possible values. The values of muted $A_M$ is selected without replacement. For the gene encoded as continuous values, $A_C$, the value is sampled from a normal distribution. This distribution is calculated from previous dominant individuals, who get more voting from the user. The evolutionary history of successful traits guides the mutation process towards optimized weights and generation of user-preferred chromosomes and individuals.

\subsection{Prompting-Free Personalized Model}
We implement the genetic prompting (optimization) model, combines with the artist model, as a real-time interactive interface where users can iteratively vote for the generated image results (n=16). Tests were conducted within the research team and with the Kandinsky expert, and in most cases, convergence was achieved within 3 to 5 iterations (likely less than 5 minutes). With this user-friendly process, users can effortlessly obtain an automated prompting model based on their own aesthetic preferences. The optimized prompting model, incorporated with the artist model, becomes the personalized model for the user. This approach effectively enables the personalization of a large text-to-image model without the need for explicit prompting. Figure \ref{fig:demo} visually illustrates the iterative genetic optimization process of the prompting model, along with the intermediate images.

\section{Discussion and Conclusion}
In this paper, we address the challenges when working with large text-to-image models, especially in the artist community. We present an artist perspective and propose an approach that heavily relies on the collective experiences of the TTI-art community and inherit established techniques to customize large model to given artistic style. We consider the prompt as an intermediary medium and draw inspiration from the practices of generative artists to control the non-deterministic occurrence of GenAI. We adapt well-established approaches from generative art, such as procedural modeling and genetic optimization to achieve automatically prompting for end users.

These techniques and algorithms enable us to exert control over large text-to-image models to generate specific artistic style. Further, it allows the user to incorporate their aesthetic preference in minimal efforts and generate personlized artistic content automatically. By incorporating expert knowledge and user preferences into the process, we aim to strike a delicate balance between complete control and total chaos, thereby achieving an optimal outcome in the spirit of generative art. By understanding and leveraging the historical evolution of generative art and its potential influence on GenAI technology, we provide insights into how to foster better collaboration with GenAI models.

\section{Limitation and Future Work} 
The presented work has limitations that need to be acknowledged. The current experiments are more artist and expert-oriented, this paper primarily focuses on the technical description but lacking user evaluation.  To further validate the approach and concept, we have planned a comprehensive user study involving participants with minimal knowledge of Kandinsky's art, aiming to collect user's perception and insights for further evaluation and improvements. The proposed approach heavily relies on expert knowledge and manual examination of literature (for semantic descriptive guideline). Future research can automate this process using natural language processing, enhancing efficiency and scalability. 



\appendix

\section{Color, Form and Composition Theory} \label{cfc}

\paragraph*{Color}
Kandinsky's color theory consists of three primary colors (i.e., \textit{Red}, \textit{Yellow}, and \textit{Blue}) and three primary tones (i.e., \textit{Black}, \textit{White}, and \textit{Grey}).  The intersection of these primary colors produces \textit{Orange}, \textit{Green}, and \textit{Violet}. He categorizes these six colors into warm and cold temperature, with red, yellow, and orange considered warm, and green, blue, and purple considered cold.  He defines the tone of a painting as light and dark; determined by the proportion of black and white in a composition. Each color represents a unique spiritual emotion and has an influence on others.\cite[pp.39-43]{Kandinsky_1977}

\paragraph*{Form}
Kandinsky presents a unique theory of tension that describes an element based on its shape, position and orientation. He begins with the concept of the \textit{Point} and derives two primary forms: Line and Plane \cite[pp.23-54]{Kandinsky_Rebay_1979}. Lines are influenced by different types of tension and can take the form of \textit{Straight Lines}, \textit{Curve Lines}, or \textit{Angular Lines} \cite[pp.55-112]{Kandinsky_Rebay_1979}. Similarly, planes are influenced by tension and can manifest as \textit{Triangles}, \textit{Squares}, or \textit{Circles} \cite[p.74]{Kandinsky_Rebay_1979}. These seven elements constitute the primary forms in Kandinsky's paintings.

\paragraph*{Composition}
Kandinsky claims interrelationships of all elements influence the composition. The composition involves the positioning of individual elements, the connections between them, and the arrangement of element clusters. Our study aims to extract easy-to-use compositions to guide the model generation and simplify Kandinsky's complex definitions into two compositional relationships: \textit{Acentric} and \textit{Centric Structure} \cite[pp.137-139]{Kandinsky_Rebay_1979}. These relationships describe the tension of elements relative to the center. In contrast, the tension of elements in relation to the composition's boundaries manifests as \textit{Inner} and \textit{External Parallel}, which is parallel to diagonal and edge of the painting respectively \cite[pp.130-131]{Kandinsky_Rebay_1979}.

\bibliographystyle{named}
\bibliography{ijcai24}

\end{document}